\documentclass[runningheads]{llncs}
\usepackage{graphicx}
\usepackage{multirow}
\usepackage{booktabs}
\begin{document}
\title{Lightweight Multi-Drone Detection and 3D-Localization via YOLO\thanks{We gratefully acknowledge the support of
NVIDIA Corporation with the donation of the Titan Xp GPU used for this
research.}}
%
%
\author{Aryan Sharma\inst{1} \and
Nitik Jain\inst{1} \thanks{The first and second authors have contributed equally.}\and
Mangal Kothari\inst{1}\orcidID{0000-0003-4713-7832 }}
\authorrunning{A. Sharma and N. Jain et al.}
%
\institute{Indian Institute of Technology Kanpur, Kanpur, India. \and
\email{\{saryan, nitik, mangal\}@iitk.ac.in}\\
}
\maketitle              
\begin{abstract}
In this work, we present and evaluate a method to perform real-time multiple drone detection and three-dimensional localization using state-of-the-art tiny-YOLOv4 object detection algorithm and stereo triangulation. Our computer vision approach eliminates the need for computationally expensive stereo matching algorithms, thereby significantly reducing the memory footprint and making it deployable on embedded systems. Our drone detection system is highly modular (with support for various detection algorithms) and capable of identifying multiple drones in a system, with real-time detection accuracy of up to 77\% with an average FPS of 332 (on Nvidia Titan Xp). We also test the complete pipeline in AirSim environment, detecting drones at a maximum distance of 8 meters, with a mean error of $23\%$ of the distance. We also release the source code for the project, with pre-trained models and the curated synthetic stereo dataset which can be found at \texttt{github.com/aryanshar/swarm-detection}
\end{abstract}
\section{Introduction}
Unmanned Aerial Vehicles (UAVs) have gained massive popularity in recent years, owing to the advancements in technology and surge in the number of use-case UAVs those include traffic management, security and surveillance, supply of essentials, disaster management, warehouse operations etc. Drones were initially a military, surveillance and security tool. But in the present era, the ecosystem around UAVs has grown into a fast growing commercial and defense markets which has in-turn drawn investments into the UAV technology, bringing down both shape, size and the costs.  Early versions of the drone were much larger, but as time progressed, they got smaller and smarter. Consequently with the development of small and agile drones, their applications have time and again raised security concerns. Their increasing use in swarm systems have also sparked another research direction in dynamic detection and localization of multiple drones in such systems, especially for counter-drone systems.  

Drone detection is essentially a subset of the widely studied object detection problem. Though the whole paradigm of object detection has witnessed use of various sensors (eg. RADARS, LiDARs etc) with various novel solutions, the real breakthrough was the use of deep-learning based methodologies for object detection and tracking. Progressively, deep learning based solutions have improved at the task of object detection, but have also grown bulkier and have relied heavily on bulky computing power. Thus, these existing methods found two roadblocks en-route to being deployed on UAVs: computing hardware and real-time inference.  

Consequently, parallel to UAV technology, development of small factor computing board and embedded computing have made it possible to deploy deep learning models on UAVs. Onboard object detection and localization has since been attracting traction. This survey \cite{applicationChallenges} summarizes the applications and associated challenges in video surveillance via drones, and highlights the recent progress and issues with the whole paradigm of drone detection and tracking. Particularly, the problem of object localization (drones in context of this work) is computationally expensive, since it relies on extracting features and matching them, a problem which makes the network bulkier and inference slower. 

In this work present a light weight computer vision pipeline for dynamic detection and localization of multiple drones. We begin with briefing about related work in this field, followed by explaining our approach. Section 4 then presents out detailed study, where we benchmark state-of-the-art object detection method YOLO and its variants on drone dataset, thereby selecting the most appropriate for our pipeline. Section 5 described our depth detection methodology in detail and concludes with the comparison of estimated depth to the ground truth. We then conclude with the summary and proposed future work which entails extending the network for detecting long range objects. 

\begin{centering}
\begin{figure*}[h!]
\includegraphics[width=\textwidth]{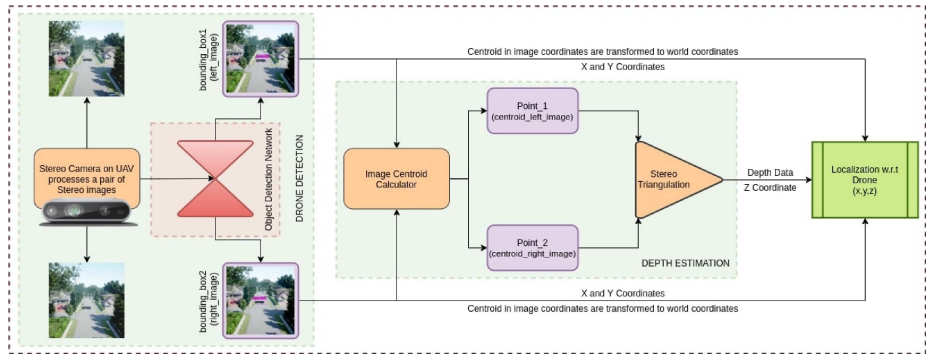}    
\caption{Process flow of our proposed pipeline for multiple drone detection and localization}
\label{fig:pipeline}
\end{figure*}
\end{centering}

\section{Related Work}

For the purpose of this section, drone detection is treated as a subset of object detection and all works have been mentioned keeping in mind their application for drone detection. Deep learning networks have increasingly been extending the generality of object detectors. In contrast to traditional methods in which each stage is individually hand-crafted and optimized by classical pipelines, deep learning networks achieve superior performance by automatically deriving each stage for feature representation and detection. 
 
In early years, video-based object detection is by extracting discriminant features such as Local Binary Pattern (LBP), Scale Invariant Feature Transform (SIFT), Histogram of Oriented Gradient (HOG) and Speeded Up Robust Features (SURF) then using these features to train the detector. Though the classical methods work in near real-time, they were soon outperformed by their learning based counterparts \cite{hulens2017autonomous}. 
 
 Deep learning methods dominate the current state-of-the-art methods when it comes to object detection, but only a selection of methods qualify for real-time applications. Initially, single stage methods such as SSD \cite{liu2016ssd} were the ones which were most qualified for real-time application, since 2-stage methods such as faster R-CNN \cite{ren2015faster} etc. were computationally expensive. As fewer proposal steps with hand-crafted features are involved in single-stage methods, they are computationally less complex than multi-state approaches that usually prioritize detection accuracy. In practice, there was active competition between multi-stage and single-stage methods for object detection tasks. In 2016, Redmond et al.~\cite{redmon2016yolo9000} surpassed SSD in both detection speed and accuracy with YOLOv2. A detailed survey which was referred while choosing detection network was done by Zhao et al.~\cite{objectdetectionsurvey}. CNNs based sensors have also been used in mobile robot localization as shown in \cite{harsh}.
  
 Since UAVs can support limited payload, significant efforts have been made to develop systems which perform computation off-board and communicate in real-time. Lee et al.~\cite{cloudCNN} demonstrated a system using Faster R-CNN,  moving the computation to an off-board computing cloud, while keeping low-level object detection and short-term navigation onboard. The research on embedded systems, capable of deploying deep convolution nets have lead to networks created specifically for UAVs, targeting high speed inference and low computational costs. Mini-YOLOv3 \cite{miniYOLO} is a real-time object specifically for embedded applications. SSD has also been deployed on drone, demonstrating fast object detectors as shown by Budiharto et al.~\cite{fastobjectdetector}. Autonomous detection and tracking of high speed ground vehicle using UAV is demonstrated in \cite{shastry2019autonomous}.  Attempts have been made to autonomously track and land a fully actuated aircraft as shown in \cite{BHARGAVAPURI2019113}.
 
The work of Hassan et al.~\cite{realtimeUAV} extends YOLOv3 for detecting UAVs in real-time. Deep learning based Real-Time Multiple-Object Detection and Tracking on embedded devices has been described in the work of  Hossain and Lee~\cite{embeded3D}. Novel approach for detection and 3D localization of drones using a hybrid motion-based object detector has been described by Srigrarom and Hoe Chew~\cite{hybriddrone}. Another approach for dynamic tracking and localization of small aerial vehicles based on computer vision has been demonstrated by Srigrarom et al.~\cite{smallaerial}. 
 
 Since vision based methods have trouble detecting multiple objects of same kind, those tend to fail in case of UAV swarms if not tuned properly. Hence, non-vision based method described an efficient strategy for accurate Detection and localization of UAV Swarms \cite{nonvisionswarm}. Another interesting approach was demonstrated by Carrio et al.~\cite{depthbased}, where they use depth maps instead of normal RGB feed to detect and localize drones. To the best of authors knowledge, we did not find work where multiple drones have been detected from single RGB images. Further, we then extend it presenting a pipeline for depth estimation which is then used for 3D localization.  

\section{Proposed Methodology}

We describe the overall process flow before explaining the two important modules in the future sections. Fig. \ref{fig:pipeline} illustrates the steps and processes involved in our pipeline. The stereo camera setup on UAV captures a stereo image which is then passed on to the object detection network. The network then outputs two images, with bounding box labels over the drone. This has been illustrated in the green box on the left of the image, which forms our drone detection module

The output from network is passed on to the image centroid calculator node, which  the respective centroid in image frame coordinates. In normal depth estimation methods, features are first identified and then the respective disparity between them gives an estimate of depth - making the overall process slow. We use the centroid of the two images as features, saving lot of computation. These centroids are then passed on to the stereo triangulation node, which outputs the estimated depth using method described in Section \ref{stereo}. This depth information is then transformed to get the $z$-coordinate of the drone. 

Having obtained the $z$-coordinate, the $x$ and $y$ coordinates are obtained by transforming the centroid coordinates from image frame to world coordinates. Hence, we obtain the tuple of $(x,y,z)$ coordinates which is used for 3D localization. 

\section{Drone Detection}

\begin{figure}[h!]
\begin{centering}
\includegraphics[width=\linewidth]{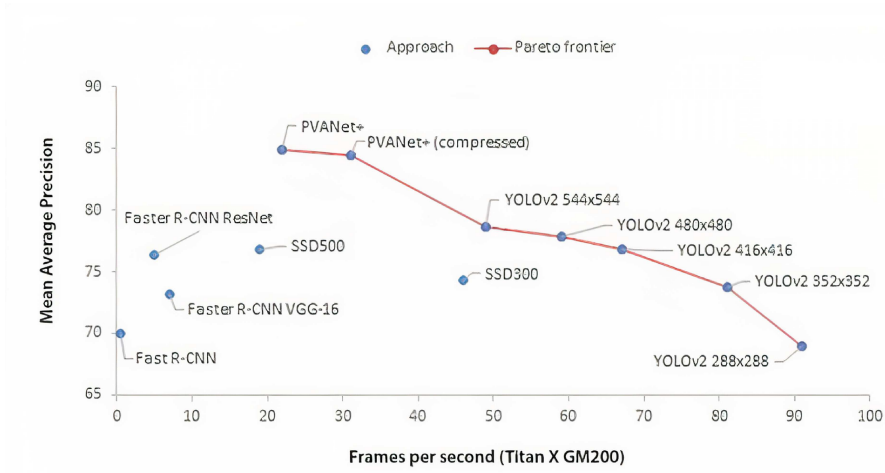}    
\caption{Comparison of different object detection algorithms tested on Pascal VOC2007 data. Image source: \texttt{https://github.com/AlexeyAB/darknet}}
\label{fig:comparison}
\end{centering}
\end{figure}

Given an image with a pixel grid representation, drone detection is the task of localizing instances of drones with bounding boxes of a certain class. Fig.~\ref{fig:comparison} is a graphical comparison of most of the previously mentioned models that plots the frame rate versus the mean average precision. The
frame rates indicated are attained using a powerful Titan X GPU and far greater than we will be able to attain, but still provide a measure for comparison among the algorithms. 

Hence, for the purpose of drone detection, a deep convolutional neural network-based model known as YOLO (You Only Look Once), essentially a state-of-the-art object detection model, is chosen and trained on a dataset of drone images. The parameters of the model have been tuned in such a way so as to better scope down the task of object detection to that of a drone detection. This work involves the training of four versions of the YOLO model namely, YOLOv3 \cite{redmon2016yolo9000}, YOLOv3-tiny \cite{tinyyolov3}, YOLOv4 \cite{bochkovskiy2020yolov4}, YOLOv4-tiny \cite{tinyyolov4}, and we compare them on the basis of some performance metrics to choose the one best suited for our problem.

\subsection{Dataset for Training}\label{dataset}

Dataset forms an integral part of the training of a neural network as the quality of a dataset directly impacts the model accuracy. In this work, the dataset has been curated from two main sources: Drone-Net \cite{dronenet} and Mendeley Drone Dataset \cite{mendeley}. In addition to this, some images have been taken from the internet and labeled manually using Labelbox annotation-tool \cite{labelbox}, so as to enrich the dataset with images containing multiple drones. Other than the images of drones, the dataset also contains images of non-drone, drone-like “negative” objects. This is done so as to avoid our model from overfitting. 

The dataset contains 5529 images along with annotated files corresponding to each image, containing parameters of bounding box such as height, width, center $x$, $y$ coordinates, and object class. The dataset is further divided into training set having 5229 images and test set having 300 images. Two different image resolutions, 450 $\times$ 280 and 1280 $\times$ 720 were tested for model training and the resolution 1280 $\times$ 720 yielded better results. Hence this resolution was chosen for every other version of YOLO that was trained.

\begin{figure}[h!]
\begin{centering}
\includegraphics[width=\linewidth]{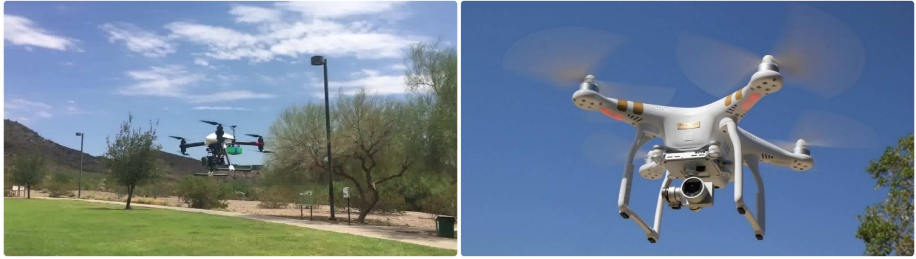}    
\caption{Sample images from Dataset used for training the detection pipeline. Higher resolution images can be found on Github repository} 
\label{fig:results}
\end{centering}
\end{figure}

\subsection{Training}

Since we were primarily comparing versions of YOLO, Darknet (\cite{darknet13}) has been used to train the models. NVIDIA TitanXP (12 GiBs vRAM) was used for training along with GPU acceleration libraries CUDA11.3 and cuDNN8.2. \par Since training heavily depends on the size of dataset, transfer learning based approach \cite{transferlearning}  has been used to counter the problem of over-fitting, which is prominent in case of smaller datasets. We begin by initializing the pretrained model given by Redmond~\cite{darknet13}. The training parameters for various models are given in Table 1.

\begin{table}[]
\centering
\caption{Parameters of CFG used for training}
\label{tab:training}
\begin{tabular}{@{}lrrrr@{}}
\toprule
\multicolumn{1}{c}{\textbf{Parameters}} &
  \multicolumn{1}{c}{\textbf{YOLOv3}} &
  \multicolumn{1}{c}{\textbf{YOLOv3-tiny}} &
  \multicolumn{1}{c}{\textbf{YOLOv4}} &
  \multicolumn{1}{c}{\textbf{YOLOv4-tiny}} \\ \midrule
Width         & 512        & 416        & 512        & 416        \\
Height        & 512        & 416        & 512        & 416        \\
Batch         & 64         & 64         & 64         & 64         \\
Subdivisions  & 16         & 2          & 32         & 16         \\
Channels      & 3          & 3          & 3          & 3          \\
Momentum      & 0.9        & 0.9        & 0.949      & 0.9        \\
Decay         & 3          & 3          & 3          & 3          \\
Learning Rate & 0.0005     & 0.0005     & 0.0005     & 0.0005     \\
max\_batch    & 0.001      & 0.001      & 0.001      & 0.001      \\
Policy        & steps      & steps      & steps      & steps      \\
Steps         & 3200, 3600 & 4480, 5040 & 3200, 3600 & 4480, 5040 \\
Scale         & 0.1, 0.1   & 0.1, 0.1   & 0.1, 0.1   & 0.1, 0.1   \\
Classes       & 1          & 1          & 1          & 1          \\ \bottomrule
\end{tabular}
\end{table}

\subsection{Results}

\begin{centering}
\begin{figure*}[h!]
\includegraphics[width=\textwidth]{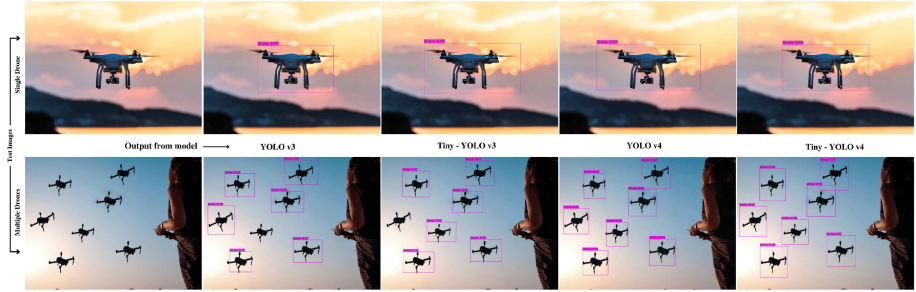}    
\caption{Results of the trained models on test images. To the left is the test image, with outputs from various model as one moves right. Top row contains image with a single drone, while bottom row is for multiple drones}
\label{fig:results}
\end{figure*}
\end{centering}

As mentioned in Section \ref{dataset}, we evaluated our models on the test set. Selected results (encompassing multiple and single drone) have been shown in Fig.~\ref{fig:results}. From the output, it can be easily seen that YOLOv4 and tiny-YOLOv4 clearly outperform YOLOv3, since YOLOv3 fails to detect all the drones due to its lower confidence values. Further, we evaluated the inference speeds, since those are of paramount importance for real-time deployment. From Table \ref{tab:results}, it can be deduced that tiny-YOLOv4 performs the best out of every model, in terms of average confidence and inference speed. Though, literature suggests that YOLOv4 must outperform every other listed models here (in terms of average confidence), but that hypothesis fails here. This is largely accredited to the the size of the dataset that the network is trained upon. Since, YOLOv4 is a very dense network, it also requires a vast dataset, both in terms of quality and quantity. Although, the choice of the network we've made is heavily based on the fact that we require a light weight network with low memory footprint and high inference speeds, so we move forward with tiny-YOLOv4. 

\begin{table}[]
\centering
\caption{Inference time \& average confidence comparison for various models on the test images in Fig. \ref{fig:results} }
\label{tab:results}
\begin{tabular}{@{}lrr@{}}
\toprule
\multicolumn{3}{c}{\textbf{Single Drone}}    \\ \midrule
\multicolumn{1}{c}{\textbf{Version}} & \textbf{Average Confidence} & \textbf{Inference Time (ms)} \\
YOLOv3            & 94\%       & 215.25      \\
tiny-YOLOv3       & 90\%       & 196.53      \\
YOLOv4            & 94\%       & 212.19      \\
tiny-YOLOv4       & 98\%       & 197.27      \\ \toprule

\multicolumn{3}{c}{\textbf{Multiple Drones}} \\ \midrule
YOLOv3            & 42\%       & 202.09      \\
tiny-YOLOv3       & 59\%       & 190.2       \\
YOLOv4            & 75\%       & 195.25      \\
tiny-YOLOv4       & 94\%       & 165.07      \\ \bottomrule
\end{tabular}
\end{table}

\subsection{Why tiny-YOLOv4~?}
The major parameters over which we have compared these four YOLO models are the confidence of prediction, inference time, and the accuracy while detecting multiple drones. On the basis of these results, we have observed the following :
\begin{itemize}
\item The newer versions of YOLO performs better overall, both in terms of accuracy and precision. 
\item Although, YOLOv4 is a much denser network, still tiny-YOLOv4 outperforms it especially in terms of the confidence in multiple object detection. 
\item YOLOv4 and tiny-YOLOv4 have much better multiple drone detection accuracy than YOLOv3 and tiny-YOLOv3
\item The Prediction Time of tiny-YOLOv4 is much better than the other YOLO versions, with also the best confidence of detection.
\end{itemize}
All of these observations motivated us to choose tiny-YOLOv4 architecture as the baseline model for the next part of the work, i.e., Depth Estimation.

\section{Depth Estimation}

The perception of depth and its estimation from a 2D image is a very challenging problem in the field of computer vision. Stereo vision consists of two identical cameras placed at a baseline distance, which allows them to take images from two distinct viewpoints. The depth in this scenario is estimated by finding the disparity of the images of the same 3D point as demonstrated by Acharyya et al.~\cite{disparity}.
The work of Yamaguchi et al.~\cite{stereomatch}, showed that one of the prominent ways to evaluate depth is through stereo-matching  of the local features in the stereo-image pair and then retrieving depth through Triangulation.
In this work, we propose an alternative approach so as to cut-down the computations involving the stereo matching process. Using tiny-YOLOv4 as the drone detection framework, we feed the stereo image frames into our trained network to retrieve the bounding box parameters of both left-camera and the right-camera images. Now, we approximate the output vector (bounding-box parameters) of the YOLO network as raw local features of the target object in the image, which makes up for the required stereo-matching and reduces the overall computation time and intensity. The approximate evaluation of depth is then done by using the x-coordinate of the centroid of the bounding box to first calculate the stereo-disparity and then use the Triangulation equation to obtain the depth.\par

\subsection{Stereo Triangulation}\label{stereo}
\begin{figure}[hbt!]
    \centering
        \includegraphics[width=0.8\linewidth]{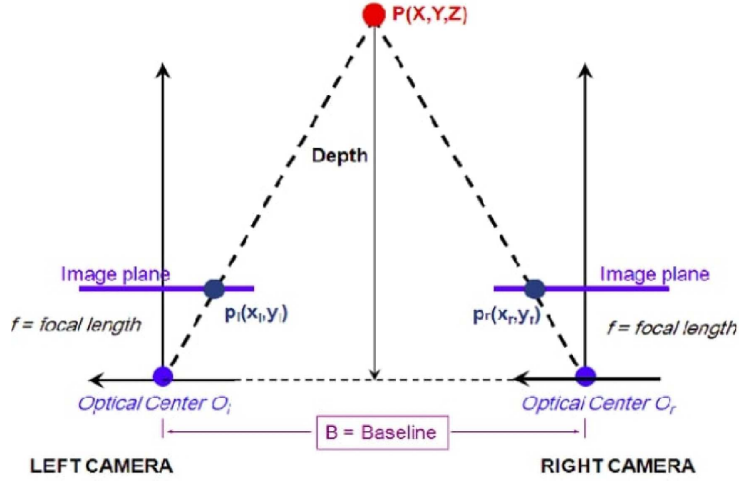}
    \caption{Stereo Triangulation scheme with parallel camera arrangement. Image source: \cite{fig-stereo}}
    \label{fig:3.2stereotriangulation}
\end{figure}

Stereo vision adds the perception of another depth dimension to a 2D digital image. The first step in the process is to evaluate the disparities between the images produced by the two cameras having a focal length $f$, and which are placed at some known baseline distance, $B$. Figure~\ref{fig:3.2stereotriangulation} shows a general arrangement for a stereo camera setup. Here, $O_{L}$ and $O_{R}$ are the optical centers for the left-camera and right-camera, respectively.

The figure also depicts the projection of a distant point object $P$ onto both the cameras. It is clear that the center of the left-camera image clearly differs from the center of the right-camera image. This is termed as parallax effect and it results in relative displacement of the respective image points when seen from different viewpoints. This displacement is termed as disparity, $\Delta  x$ which is given by:
\begin{equation}
\Delta  x = x_{L} - x_{R}
\end{equation}
From the property of similar triangles, the parameters like baseline $B$, focal length $f$, disparity $\Delta x$, and the depth $Z$, can be formulated as equality of ratios,
\begin{equation}
\frac{Z}{B} = \frac{f}{\Delta x}
\end{equation}
Obtaining the depth from this equation, requires rearranging these terms,
\begin{equation}
Z = \frac{f \times B}{\Delta x}
\label{eq:triangulation}
\end{equation}
It is evident from equation (\ref{eq:triangulation}) that if the disparity term is kept constant, then decreasing the baseline B also decreases the depth $Z$. Thus, it is suggested that to obtain depth of objects that are far away, the baseline should be kept large.

\subsection{Stereo Dataset formation using AirSim}\label{AA}

A very prominent obstacle faced throughout was the lack of dataset containing drone images, and while also considering that physically creating a dataset by flying the drones and capturing its images from different viewpoints is a very time-intensive task, inspired to create the stereo-dataset in a virtual environment, which mimics reality, by using AirSim \cite{airsim} and automating the process of creating the dataset by writing a Python script for the simulation process. AirSim exhibits certain APIs which can be used to interact with the vehicle in the simulation programmatically to retrieve images, get state, control the vehicle, etc. \par
\begin{figure}[hbt!]
\centering
\includegraphics[width=\linewidth]{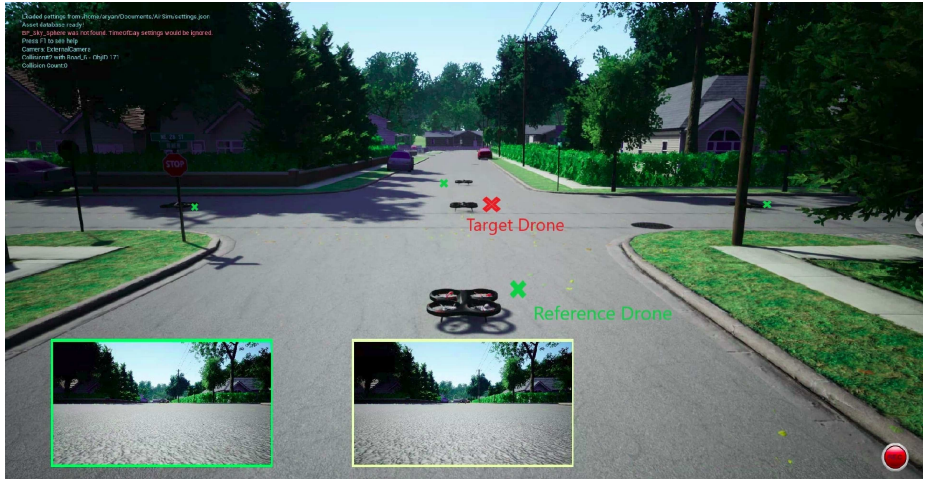}
\caption{Custom AirSim enviroment setup used for simulation. Red-cross denotes the Target Drone and Green-cross denotes the Reference Drones}
\label{airsim}\end{figure}

In order to automate the simulation process of capturing the stereo images from different viewpoints for enriching the dataset, the python script is run simultaneously with our custom AirSim environment. The arrangement in the simulation environment consists of two sets of drones (as shown in Fig.~\ref{airsim}), one are the Target Drones (whose image is being captured) that are made to fly up and hover at a particular height, and the other ones include a set of four reference drones (those who capture images of target drones). These reference drones are placed equidistant from each other, having the target drones at the center and these are made to fly such that they capture stereo images of these target drones from different viewpoints.

\begin{figure}[hbt!]
\centering
\includegraphics[width=\linewidth]{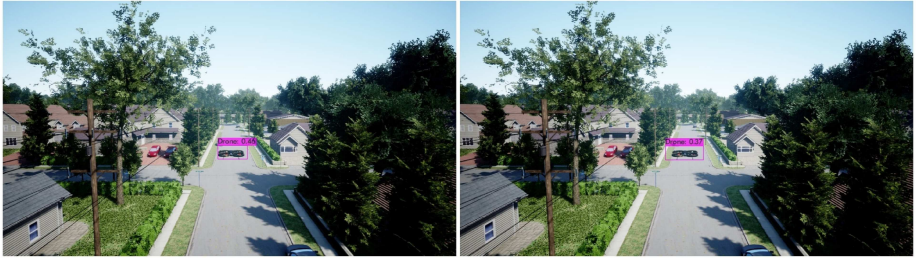}
\caption{Sample stereo-image pair of a scene after being fed through tiny-YOLOv4 network} \label{fig12}
\end{figure}
\subsection{Evaluation of Approximate Depth}

Equation~(\ref{eq:triangulation}) suggests that given the camera parameters, namely focal length ($f$) and baseline distance ($B$), if we are able to obtain the disparity between the images, the parameter of depth could then be evaluated. To do so, we feed the stereo-image pair into the trained tiny-YOLOv4 network to extract the predicted bounding box parameters. Since the stereo-camera platform that we have taken is a parallel camera arrangement, the disparity can be approximated as the difference in the $x$-coordinate of the target drone’s projection in both the left-camera and right-camera image, respectively. Now, the $x$-coordinate of the target drone’s projection in both the cameras is approximated as the $x$-coordinate of the centroid of the bounding-box obtained after feeding the images into the trained tiny-YOLOv4 network (see Figure \ref{fig12}). This establishes the requirement for obtaining the depth parameter. Say, $Cx_{LeftCam}$ and $Cx_{RightCam}$ denote the $x$-coordinate of the centroid of the bounding box parameter in the left-camera image and the right-camera image respectively, then disparity is given as:
\begin{equation}
\Delta  x_{img} = Cx_{LeftCam} - Cx_{RightCam}
\end{equation}

\subsection{Formulation for Comparison}

To establish a comparison of our approximate Depth-Detection technique with the Ground-Truth, we first generate a stereo dataset of 50 images using the simulation script as mentioned in Section~\ref{AA}. The camera parameters used for generating these images were,
Baseline, $B$ = 1.2 $m$ and focal length, $f$ = 1.2 $m$\\
Since, the bounding-box parameters given as output vector by the tiny-YOLOv4 network is in terms of the pixel no., the disparity $\Delta  x_{img}$ value obtained is terms of pixel number. Thus, making appropriate transformations using the conversion 1 $pixel$ = $2.65 \times 10^{-4}$ $m$ , we get
\begin{equation}
    Z = \frac{B \times f}{\Delta x_{img} (in\;pixels)} = \frac{9070.86}{\Delta x_{img}}(\;m)
\label{eq:finaldepth}
\end{equation}

\subsection{Comparison with Ground-Truth}
As we have the required relation for evaluating the depth of a target drone, we begin to compare its performance. To evaluate the performance of our depth estimation model, the ground truth values of the target drone and the reference drone were recorded throughout the process of simulation while capturing the dataset of 50 stereo-pair images. Out of the 50 pairs, we compute the values of depth for a sample size of 8 stereo-image pairs using Eq.~(\ref{eq:finaldepth}) and the results are reported in Table~\ref{tab:comparison} below. 

\begin{table}[htbp]
    \caption{Comparison of Estimated Z-depth and Ground Truth Depth}
    \label{tab:comparison}
    \centering
    \setlength{\tabcolsep}{10pt}
    \renewcommand{\arraystretch}{1.3}
    \begin{tabular}{c c c c c}
    Sample & Disparity & Z-Depth & Ground-Truth & Error \\
    No. & (\textit{pixel}) & (\textit{m}) & (\textit{m}) & \%\\ \hline
    1 & 1412 & 6.42 & 5.71 & 12.32 \%\\ 
    2 & 1274 & 7.12 & 5.93 & 19.97 \%\\ 
    3 & 1173 & 7.73 & 6.37 & 21.24 \%\\ 
    4 & 1104 & 8.21 & 6.54 & 25.62 \%\\ 
    5 & 1089 & 8.33 & 6.73 & 23.75 \%\\ 
    6 & 1028 & 8.82 & 6.93 & 27.25 \%\\ 
    7 & 898 & 10.10 & 7.59 & 32.95 \%\\ 
    8 & 963 & 9.42 & 7.70 & 22.27 \%\\  \hline
    \end{tabular}
\end{table}


It is observed from the table that the proposed depth estimation model gives close to accurate measurement of the object depth. It is also observed that the error in the measurement of depth from our formulated model increases with the depth. This tells us that the model is able to deliver good results when the target drone is in a close proximity.

\section{Conclusion}
We have presented and evaluated our approach for real-time multiple drone detection and localization in simulation environment. The pipeline is modular with support for various object detection algorithms depending on frame rate, YOLO and its variants have been evaluated in this work. The modern, neural net based tiny-YOLO v4 algorithm attains higher frame rates and detection accuracy results than leading CPU based algorithms, and coupled with our classical stereo triangulation based depth estimation module, can be used for 3D localization. Further, we also release the pre-trained models, source code for simulation and the custom stereo dataset for further developments. Since traditional feature matching methods tend to deteriorate as images have more negative space, our method circumvents the problem since we use centroid as a feature for stereo triangulation. Hence as future improvements, we plan to extend our work for achieving long range drone detection using UAV-Yolo \cite{uavyolo}. 

\bibliographystyle{splncs04}
\bibliography{drone_det}

\begin{thebibliography}{10}
\providecommand{\url}[1]{\texttt{#1}}
\providecommand{\urlprefix}{URL }
\providecommand{\doi}[1]{https://doi.org/#1}

\bibitem{disparity}
Acharyya, A., Hudson, D., Chen, K.W., Feng, T., Kan, C.Y., Nguyen, T.: Depth
  estimation from focus and disparity. In: 2016 IEEE International Conference
  on Image Processing (ICIP). pp. 3444--3448 (2016).
  \doi{10.1109/ICIP.2016.7532999}

\bibitem{tinyyolov3}
Adarsh, P., Rathi, P., Kumar, M.: Yolo v3-tiny: Object detection and
  recognition using one stage improved model. In: 2020 6th International
  Conference on Advanced Computing and Communication Systems (ICACCS). pp.
  687--694 (2020). \doi{10.1109/ICACCS48705.2020.9074315}

\bibitem{mendeley}
Aksoy, M.a.: Drone dataset: Amateur unmanned air vehicle detection (Nov 2019),
  \url{https://data.mendeley.com/datasets/zcsj2g2m4c}

\bibitem{BHARGAVAPURI2019113}
Bhargavapuri, M., Shastry, A.K., Sinha, H., Sahoo, S.R., Kothari, M.:
  Vision-based autonomous tracking and landing of a fully-actuated rotorcraft.
  Control Engineering Practice  \textbf{89},  113--129 (2019).
  \doi{https://doi.org/10.1016/j.conengprac.2019.05.015},
  \url{https://www.sciencedirect.com/science/article/pii/S0967066118306415}

\bibitem{bochkovskiy2020yolov4}
Bochkovskiy, A., Wang, C.Y., Liao, H.Y.M.: Yolov4: Optimal speed and accuracy
  of object detection (2020)

\bibitem{fastobjectdetector}
Budiharto, W., Gunawan, A.A.S., Suroso, J.S., Chowanda, A., Patrik, A., Utama,
  G.: Fast object detection for quadcopter drone using deep learning. In: 2018
  3rd International Conference on Computer and Communication Systems (ICCCS).
  pp. 192--195 (2018). \doi{10.1109/CCOMS.2018.8463284}

\bibitem{depthbased}
Carrio, A., Tordesillas, J., Vemprala, S., Saripalli, S., Campoy, P., How,
  J.P.: Onboard detection and localization of drones using depth maps. IEEE
  Access  \textbf{8},  30480--30490 (2020). \doi{10.1109/ACCESS.2020.2971938}

\bibitem{dronenet}
Chuanenlin: chuanenlin/drone-net: 2664 images of drones, labeled, with trained
  yolo weights. example project for my article "tutorial: Build your custom
  real-time object classifier" on medium. (Aug 2018),
  \url{https://github.com/chuanenlin/drone-net}

\bibitem{applicationChallenges}
Dilshad, N., Hwang, J., Song, J., Sung, N.: Applications and challenges in
  video surveillance via drone: A brief survey. In: 2020 International
  Conference on Information and Communication Technology Convergence (ICTC).
  pp. 728--732 (2020). \doi{10.1109/ICTC49870.2020.9289536}

\bibitem{fig-stereo}
Fahmy, A.: Stereo vision based depth estimation algorithm in uncalibrated
  rectification (2013)

\bibitem{realtimeUAV}
Hassan, S.A., Rahim, T., Shin, S.Y.: Real-time uav detection based on deep
  learning network. In: 2019 International Conference on Information and
  Communication Technology Convergence (ICTC). pp. 630--632 (2019).
  \doi{10.1109/ICTC46691.2019.8939564}

\bibitem{embeded3D}
Hossain, S., Lee, D.j.: Deep learning-based real-time multiple-object detection
  and tracking from aerial imagery via a flying robot with gpu-based embedded
  devices. Sensors  \textbf{19}(15) (2019). \doi{10.3390/s19153371},
  \url{https://www.mdpi.com/1424-8220/19/15/3371}

\bibitem{hulens2017autonomous}
Hulens, D., Goedem{\'e}, T.: Autonomous flying cameraman with embedded person
  detection and tracking while applying cinematographic rules. In: 2017 14th
  Conference on Computer and Robot Vision (CRV). pp. 56--63. IEEE (2017)

\bibitem{labelbox}
Labelbox: Labelbox/labelbox: Labelbox is the fastest way to annotate data to
  build and ship computer vision applications.,
  \url{https://github.com/Labelbox/labelbox}

\bibitem{cloudCNN}
Lee, J., Wang, J., Crandall, D., Šabanović, S., Fox, G.: Real-time,
  cloud-based object detection for unmanned aerial vehicles. In: 2017 First
  IEEE International Conference on Robotic Computing (IRC). pp. 36--43 (2017).
  \doi{10.1109/IRC.2017.77}

\bibitem{uavyolo}
Liu, M., Wang, X., Zhou, A., Fu, X., Ma, Y., Piao, C.: Uav-yolo: Small object
  detection on unmanned aerial vehicle perspective. Sensors  \textbf{20}(8)
  (2020). \doi{10.3390/s20082238},
  \url{https://www.mdpi.com/1424-8220/20/8/2238}

\bibitem{liu2016ssd}
Liu, W., Anguelov, D., Erhan, D., Szegedy, C., Reed, S., Fu, C.Y., Berg, A.C.:
  Ssd: Single shot multibox detector. In: European conference on computer
  vision. pp. 21--37. Springer (2016)

\bibitem{miniYOLO}
Mao, Q.C., Sun, H.M., Liu, Y.B., Jia, R.S.: Mini-yolov3: Real-time object
  detector for embedded applications. IEEE Access  \textbf{7},  133529--133538
  (2019). \doi{10.1109/ACCESS.2019.2941547}

\bibitem{transferlearning}
Pan, S.J., Yang, Q.: A survey on transfer learning. IEEE Transactions on
  Knowledge and Data Engineering  \textbf{22}(10),  1345--1359 (2010).
  \doi{10.1109/TKDE.2009.191}

\bibitem{darknet13}
Redmon, J.: Darknet: Open source neural networks in c.
  \url{http://pjreddie.com/darknet/} (2013--2016)

\bibitem{redmon2016yolo9000}
Redmon, J., Farhadi, A.: Yolo9000: Better, faster, stronger (2016)

\bibitem{ren2015faster}
Ren, S., He, K., Girshick, R., Sun, J.: Faster r-cnn: Towards real-time object
  detection with region proposal networks. Advances in neural information
  processing systems  \textbf{28},  91--99 (2015)

\bibitem{airsim}
Shah, S., Dey, D., Lovett, C., Kapoor, A.: Airsim: High-fidelity visual and
  physical simulation for autonomous vehicles (2017)

\bibitem{shastry2019autonomous}
Shastry, A.K., Sinha, H., Kothari, M.: Autonomous detection and tracking of a
  high-speed ground vehicle using a quadrotor uav. In: AIAA Scitech 2019 Forum.
  p.~1188 (2019)

\bibitem{harsh}
Sinha, H., Patrikar, J., Dhekane, E.G., Pandey, G., Kothari, M.: Convolutional
  neural network based sensors for mobile robot relocalization. In: 2018 23rd
  International Conference on Methods Models in Automation Robotics (MMAR). pp.
  774--779 (2018). \doi{10.1109/MMAR.2018.8485921}

\bibitem{hybriddrone}
Srigrarom, S., Hoe~Chew, K.: Hybrid motion-based object detection for detecting
  and tracking of small and fast moving drones. In: 2020 International
  Conference on Unmanned Aircraft Systems (ICUAS). pp. 615--621 (2020).
  \doi{10.1109/ICUAS48674.2020.9213912}

\bibitem{smallaerial}
Srigrarom, S., Lee, S.M., Lee, M., Shaohui, F., Ratsamee, P.: An integrated
  vision-based detection-tracking-estimation system for dynamic localization of
  small aerial vehicles. In: 2020 5th International Conference on Control and
  Robotics Engineering (ICCRE). pp. 152--158 (2020).
  \doi{10.1109/ICCRE49379.2020.9096259}

\bibitem{stereomatch}
Yamaguchi, K., McAllester, D., Urtasun, R.: Efficient joint segmentation,
  occlusion labeling, stereo and flow estimation. In: Fleet, D., Pajdla, T.,
  Schiele, B., Tuytelaars, T. (eds.) Computer Vision -- ECCV 2014. pp.
  756--771. Springer International Publishing, Cham (2014)

\bibitem{objectdetectionsurvey}
Zhao, Z.Q., Zheng, P., Xu, S.T., Wu, X.: Object detection with deep learning: A
  review. IEEE Transactions on Neural Networks and Learning Systems
  \textbf{30}(11),  3212--3232 (2019). \doi{10.1109/TNNLS.2018.2876865}

\bibitem{nonvisionswarm}
Zheng, J., Chen, R., Yang, T., Liu, X., Liu, H., Su, T., Wan, L.: An efficient
  strategy for accurate detection and localization of uav swarms. IEEE Internet
  of Things Journal pp.~1--1 (2021). \doi{10.1109/JIOT.2021.3064376}

\bibitem{tinyyolov4}
Zhu, D., Xu, G., Zhou, J., Di, E., Li, M.: Object detection in complex road
  scenarios: Improved yolov4-tiny algorithm. In: 2021 2nd Information
  Communication Technologies Conference (ICTC). pp. 75--80 (2021).
  \doi{10.1109/ICTC51749.2021.9441643}

\end{thebibliography}
\end{document}